\documentclass[10pt,twocolumn]{article}
\usepackage{times}
\usepackage{graphicx}
\usepackage{amsmath}
\usepackage{amssymb}
\usepackage[margin=20truemm]{geometry}

\begin{document}

\title{Joint Feature Distribution Alignment Learning\\for NIR-VIS and VIS-VIS Face Recognition}

\author{Takaya Miyamoto$^1$, Hiroshi Hashimoto$^1$, Akihiro Hayasaka$^1$, Akinori F. Ebihara$^1$, Hitoshi Imaoka$^2$ \\
$^1$NEC Biometrics Research Laboratories, $^2$NEC Corporation \\ \ 
}
\date{}
\maketitle

\begin{abstract}
   Face recognition for visible light (VIS) images achieve high accuracy thanks to the recent development of deep learning.
   However, heterogeneous face recognition (HFR), which is a face matching in different domains, is still a difficult task due to the domain discrepancy and lack of large HFR dataset.
   Several methods have attempted to reduce the domain discrepancy by means of fine-tuning, which causes significant degradation of the performance in the VIS domain because it loses the highly discriminative VIS representation. 
   To overcome this problem, we propose joint feature distribution alignment learning (JFDAL) which is a joint learning approach utilizing knowledge distillation.
   It enables us to achieve high HFR performance with retaining the original performance for the VIS domain. 
   Extensive experiments demonstrate that our proposed method delivers statistically significantly better performances compared with the conventional fine-tuning approach on a public HFR dataset Oulu-CASIA NIR\&VIS and popular verification datasets in VIS domain such as FLW, CFP, AgeDB. 
   Furthermore, comparative experiments with existing state-of-the-art HFR methods show that our method achieves a comparable HFR performance on the Oulu-CASIA NIR\&VIS dataset with less degradation of VIS performance.
\end{abstract}

\section{Introduction}

\begin{figure}[t]
   \begin{center}
      \includegraphics[width=0.9\linewidth]{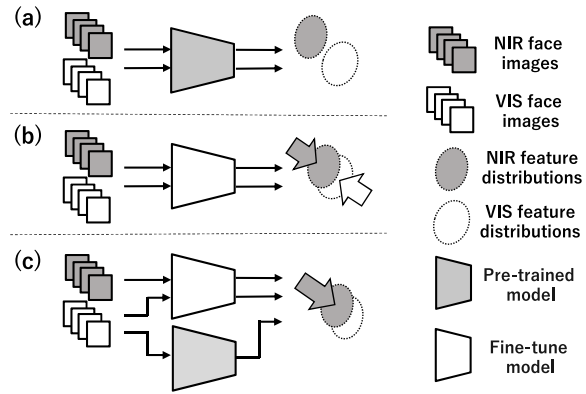}
   \end{center}
   \caption{Schematic diagrams for intra-class feature distributions in the feature space.
   (a) The feature distributions of visible light (VIS) images and near-infrared light (NIR) images have a gap due to a large domain discrepancy.
   (b) Conventional HFR approaches fine-tune the pre-trained model by utilizing a HFR dataset to reduce the domain discrepancy, but the highly discriminative feature for the VIS domain changes and recognition performance for VIS face image degrades.
   (c) Our proposed method aligns the NIR-VIS feature distributions and retains the VIS feature distribution from its original position simultaneously.}
   \label{fig:intro_cartoon}
\end{figure}

Face recognition enables non-contact and un-constrained biometric authentication system and has been widely used in real applications such as security surveillance and access control.
The performance of face recognition for visible light (VIS) images has been rapidly improving thanks to the development of deep learning \cite{LeNet,CosF,ArcF} and large-scale public VIS face datasets such as CASIA-webface \cite{CASIA}, VGGface2 \cite{VGG2}, and MS-Celeb 1M \cite{MS1M}.
In the real applications, however, various image domains are utilized depending on the usage.
For example, a near-infrared light (NIR) image is effective for recognition in a night-time scene.
In such a case, the system carries out matching between NIR and VIS face images because a typical face recognition system requires users to enroll with their VIS images.
Such a cross-domain face-matching task is referred to as heterogeneous face recognition (HFR) which has been widely investigated \cite{HFRsurvey,WCNN,DVR,DVG,DVGFace} because it is still a difficult task due to a large domain discrepancy.

Conventional HFR methods utilize models pre-trained with large-scale VIS face images to ensure the performance \cite{HFRsurvey}.
The pre-trained model has a gap between NIR and VIS feature distribution due to a large domain discrepancy (shown in Fig.~\ref{fig:intro_cartoon}~(a)), which increases the intra-class distance and degrades the HFR performance.
Thus a fine-tuning (FT) approach is utilized to reduce the intra-class domain discrepancy by using a HFR dataset including cross-domain images of the same person, as shown in Fig.~\ref{fig:intro_cartoon} (b).
Several works \cite{WCNN, DVR, DVG, DVGFace} focused on the development of loss funcition for the FT to minimize the discrepancy of the feature distributions in different domains.
Although the FT approach achieves a state-of-the-art HFR performance, it causes a significant performance degradation of the source task (i.e., the VIS performance) which is related to the catastrophic forgetting \cite{CFsurvey}.
Achieving a high HFR performance without degrading the VIS performance is therefore a challenging problem.
To open up a greater range of possible application scenarios for a single face recognition system, it is necessary to achieve high VIS-VIS and NIR-VIS performances simultaneously with a single model.

If both a source and a target dataset can be utilized simultaneously, joint training (JT) \cite{MultiTask} is a simple but effective approach to ensure the recognition performance of both the source and target tasks.
In such a case, it is expected that the recognition performance strongly depends on size of each training dataset.
Thus, in the case of a very small target dataset, simple JT may not achieve a high performance for the target task due to fewer appearances of target samples in a training batch, which is observed in the HFR task (e.g., see Table~5 in \cite{DVGFace}).
In fact, since the collection of a large number of heterogeneous face images is expensive and time-consuming, typical HFR datasets have smaller number of face images compared to large-scale VIS datasets.
For example, the Oulu-CASIA NIR\&VIS dataset \cite{OuluCASIA}, one of the publicly available HFR datasets, contains about 65k face images, while the MS-Celeb 1M dataset has about 10 million face images.
There is also a problem in that up-sampling of the HFR data may causes strong over-fitting and degrades the VIS performance.

This paper aims to overcome the above problem, which is to achieve high performance for both VIS-VIS and NIR-VIS face recognition simultaneously.
To this end, we propose joint feature distribution alignment learning (JFDAL) which provides an effective trade-off between FT and JT.
JFDAL consists of two parts: cross-domain feature distribution alignment Learning (CFDAL) and source-domain feature distribution alignment learning (SFDAL). 
CFDAL imposes the feature alignment between different domains to minimize the domain discrepancy in the same way as fine-tuning-based methods, while SFDAL constrains source domain features in a knowledge distillation manner.
Applying only CFDAL, namely FT as depicted in Fig.~\ref{fig:intro_cartoon}~(b), will change not only NIR features but also VIS features. 
This change may lose a pre-trained model's highly discriminative power, which leading a degradation of VIS performance.
By jointly utilizing CFDAL and SFDAL, our proposed JFDAL aligns the NIR-VIS feature distributions and retains the VIS feature distribution from its original position simultaneously, as shown in Fig.~\ref{fig:intro_cartoon} (c). 
As a result, the trained model achieves a high HFR performance with less degradation of the VIS performance.

Our contributions are threefold:
\begin{itemize}
   \item We introduce a new task on the HFR for achieving a high HFR performance without degrading the VIS performance on a single model.
   \item We propose joint feature distribution alignment learning (JFDAL), which is a learning method to align the NIR-VIS feature distributions while retaining the VIS feature distribution from its original position.
   \item We extensively evaluate our approach on LFW, CALFW, CPLFW, CFP-FP, AgeDB-30 for the VIS face dataset and on Oulu-CASIA NIR\&VIS for the small HFR dataset.
   Our approach achieves a superior performance to the conventional joint training and fine-tuning approaches.
\end{itemize}

\begin{figure*}
   \begin{center}
      \includegraphics[width=0.9\linewidth]{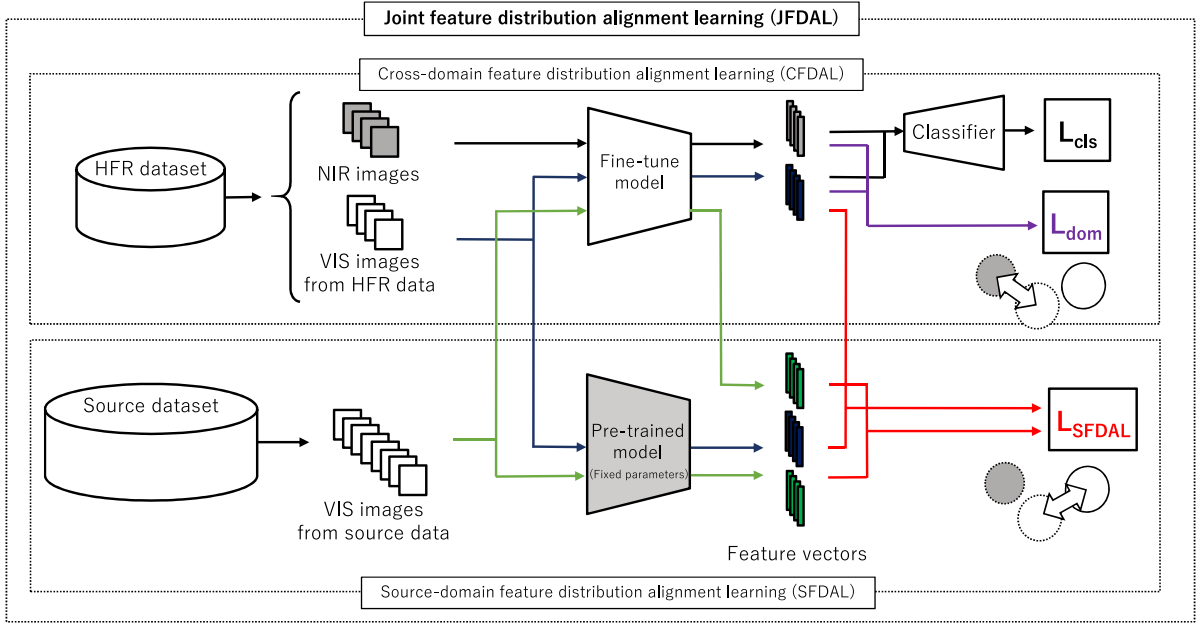}
   \end{center}
   \caption{Overview of proposed joint feature distribution alignment learning (JFDAL). 
   JFDAL consists of two parts: cross-domain feature distribution alignment learning (CFDAL) and source-domain feature distribution alignment learning (SFDAL).
   CFDAL minimizes a gap between the feature distribution of NIR face images (gray circle with dotted line) and that of VIS face images (while circle with dotted line).
   SFDAL utilizes VIS face images in both HFR dataset and large-scale VIS dataset and constrains VIS features to keep the well-discriminated distributions (white circle with solid line) in a knowledge distillation manner.
   Note that CFDAL uses only the fine-tune model, while SFDAL uses both the fine-tune and pre-trained models.}
   \label{fig:method}
\end{figure*}

\section{Related works}

Our proposed JFDAL takes inspiration from two separate research areas: heterogeneous face recognition and incremental learning.
These areas have been investigated extensively, and we introduce them in the following two subsections.

\subsection{Heterogeneous face recognition}

Heterogeneous face recognition (HFR) refers to the matching of face images that belong to different domains \cite{HFRsurvey}.
The main objective of HFR is to reduce the domain discrepancy in the feature space so as to improve the verification performance. 
Typically, there are three ways of doing this: (i) synthesizing a face image to a different domain \cite{Fsynth_1, Fsynth_2, Fsynth_3}, (ii) projecting features to a common feature sub-space \cite{PrjComm_1, PrjComm_2, PrjComm_3, PrjComm_4}, and (iii) training a model to extract the domain-invariant feature \cite{DinvF_1, DVR, MMDL}.
\cite{WCNN} used a Siamese neural network to learn domain-invariant features by minimizing the Wasserstein distance between the distributions of NIR and VIS features.
\cite{DVR} proposed disentangled variational representation (DVR) for cross-domain matching, where the variational lower bound is used to estimate the posterior and optimize the disentangled latent variable space.
\cite{DVG} introduced the dual variational generation (DVG) method, which promotes the inter-class diversity by generating massive new pairs of VIS and NIR images from noise. 
The synthesized pairs are then used to fine-tune the pre-trained models to reduce domain discrepancy by a simple pairwise distance loss.
\cite{DVGFace} developed an improved version of DVG in which unpaired VIS images from the large-scale VIS dataset are utilized to enrich the identity diversity of the generated images.
These methods focus only on cross-domain discrepancy, and do not take an VIS feature distribution into account.

\subsection{Incremental learning}

Training a pre-trained model without forgetting a source task, also referred as the incremental learning, has been a long-standing problem \cite{CFsurvey,LwF,iCaRL}.
Li and Hoiem proposed LwF \cite{LwF}, which introduces knowledge distillation \cite{Distil} to preserve the knowledge of a pre-trained model. 
They used only samples of the new class and implemented a distillation loss function to preserve the old class information.
Rebuffi et al. proposed iCaRL \cite{iCaRL}, which allows the use of a few samples of the old class.
They proposed a method to select a small number of exemplars from each old class to preserve old knowledge effectively. 
For the face recognition task, Liu et al. proposed FAwF \cite{FAwF}, a fast domain adapting method with less degradation of the source domain performance.
In these works, the problem of ``forgetting'' have been discussed in term of classification task. 
Since HFR is a feature embedding task, effectiveness of these methods for HFR is unkown. 

\section{Proposed method}

In this section, we introduce our proposed method, JFDAL.
The overview of JFDAL is shown in Fig.~\ref{fig:method}.
The two components of our method, CFDAL and SFDAL, are presented step-by-step in the next subsections.

Throughout this section, we denote the feature extraction process as $X^i_d = F(I^i_d, \Theta)$, where $I^i_d$ denotes a input face image for identity $i$ in $d$ domain ($d=\mathrm{N}$ for NIR domain and $d=\mathrm{V}$ for VIS domain) sampled from dataset $\mathcal{D}$, $X^i_d$ stands for the extracted feature vector, $F(\cdot)$ is a feature extractor parametrized by $\Theta$.
We also define $\Theta_{\mathrm{init}}$ as the parameters from the model pre-trained on a large-scale VIS dataset.

\subsection{Cross-domain feature distribution alignment learning}

The first component of our method, CFDAL, is utilized for training to reduce the distance of feature distributions in different domains.
In each training iteration, we sample $N$ face images from the HFR dataset as the training batch $\mathcal{B}$ ($|\mathcal{B}| = N$).
Inspired by conventional HFR methods \cite{DVG, DVGFace}, we implement two objective functions: $L_{\mathrm{dom}}$ which measures the distance of the two feature distributions in different domains, and $L_{\mathrm{cls}}$, which is a face classification loss function.
The goal of CFDAL is to minimize $L_{\mathrm{CFDAL}} = L_{\mathrm{cls}} + \lambda L_{\mathrm{dom}}$, where $\lambda \in [0,\infty)$ is a hyperparameter.

For the metric of the two feature distributions, we utilize the maximum mean discrepancy (MMD) \cite{MMD, MMDcnn} for each identity.
Assuming there are $M$ identities in the training batch sampled from the HFR dataset, $L_{\mathrm{dom}}$ is given by
\begin{eqnarray}
   L_{\mathrm{dom}} &=& \frac{1}{M} \sum_i \left|\left| \mu^i_\mathrm{N} - \mu^i_\mathrm{V} \right|\right|, \\
   \mu^i_d &=& \frac{1}{|\mathcal{B}_d|} \sum_{I^i_d \in \mathcal{B}_d} F(I^i_d, \Theta) \hspace{0.5cm} (d\in\{\mathrm{N},\mathrm{V}\}) \nonumber
\end{eqnarray}
where $\mathcal{B}_\mathrm{N}$, $\mathcal{B}_\mathrm{V}$ are the NIR, VIS domain subset in the training batch, respectively.

For the classification loss $L_{\mathrm{cls}}$, we utilize a angular-margin softmax loss \cite{CosF} with class labels as in \cite{FAwF}.

\subsection{Source-domain feature distribution alignment learning}

SFDAL, the second component of our method, is utilized for training to retain the VIS feature distributions from their original points.
To this end, we use the pre-trained model with fixed parameters as a guidance model to indicate the VIS feature distributions \cite{TeaNet}.
Treating the feature distributions extracted from the pre-trained model as a teacher, and to minimize the distances from the distributions of the training model, our method constrains the model to keep the VIS feature distribution during the training.

In SFDAL, we sample $N^\prime$ face images from the large-scale VIS dataset as the training batch $\mathcal{B}^\prime$ ($|\mathcal{B}^\prime|=N^\prime$).
For an objective, we choose the minimization of an L2-distance between the features extracted from the parameter-fixed pre-trained model $F(I^i_d, \Theta_{\mathrm{init}})$ and the training model $F(I^i_d, \Theta)$.
Further, we estimate the distance of the VIS images from both the large-scale VIS dataset and the HFR dataset.
Thus, the loss function to be minimized is given by
\begin{eqnarray}
   L_{\mathrm{SFDAL}} = \frac{1}{|\bar{\mathcal{B}}|} \sum_{I^i_V \in \bar{\mathcal{B}}} \left|\left| F(I^i_V, \Theta_{\mathrm{init}}) - F(I^i_V, \Theta) \right|\right|,
\end{eqnarray}
where $\bar{\mathcal{B}} = \mathcal{B}^\prime+\mathcal{B}_\mathrm{V}$.

\subsection{Joint feature distribution alignment learning}

By jointly applying CFDAL and SFDAL, the model is trained so that it reduces the domain discrepancy and retains the VIS domain feature distributions from the pre-trained model, simultaneously.
We sample $N$ face images from the HFR dataset and $N^\prime$ face images from the large-scale VIS dataset in each iteration and then calculate both loss functions $L_{\mathrm{CFDAL}}$ and $L_{\mathrm{SFDAL}}$.
The total loss in JFDAL is given by
\begin{eqnarray}
   L_{\mathrm{tot}} = \alpha L_{\mathrm{CFDAL}} + (1-\alpha) L_{\mathrm{SFDAL}}, \label{eq:Ltot}
\end{eqnarray}
where $\alpha \in [0,1]$ stands for a trade-off parameter that balances the effect of CFDAL and SFDAL.
When we choose $\alpha = 1$, JFDAL is reduced to a fine-tuning with only the HFR dataset.
The effect of $\alpha$ is discussed in Sec.\ref{sec:hparam_analysis}.

\section{Experiments}

\begin{table*}[t]
   \caption{Accuracy [\%] on various VIS verification datasets for three baseline models (VIS-only, joint training, and fine-tuning) and our proposed method. Parentheses indicate standard deviations.}
   \begin{center}
      \begin{tabular}{|l|c|c|c|c|c|c||c|}
         \hline
         Method & LFW & CALFW & CPLFW & CFP-FP & CFP-FF & AgeDB-30 & Average \\
         \hline\hline
         VIS-only & 99.74 (0.01) & 94.93 (0.05) & 90.73 (0.12) & 95.44 (0.10) & 99.71 (0.03) & 96.60 (0.11) & 96.19 (0.02) \\
         Joint training & 99.74 (0.02) & 94.93 (0.07) & 90.62 (0.07) & 95.42 (0.03) & 99.71 (0.02) & 96.61 (0.08) & 96.17 (0.01) \\
         Fine-tuning & 99.63 (0.01) & 94.70 (0.09) & 90.33 (0.14) & 95.07 (0.10) & 99.58 (0.01) & 95.67 (0.13) & 95.83 (0.07) \\
         Proposed & 99.68 (0.00) & 95.17 (0.02) & 90.89 (0.04) & 95.12 (0.03) & 99.58 (0.01) & 96.21 (0.04) & 96.11 (0.01) \\
         \hline
         \end{tabular}
      \end{center}
   \label{tab:results_1_vis}
\end{table*}

\begin{figure*}[t]
   \begin{center}
      \includegraphics[width=0.9\linewidth]{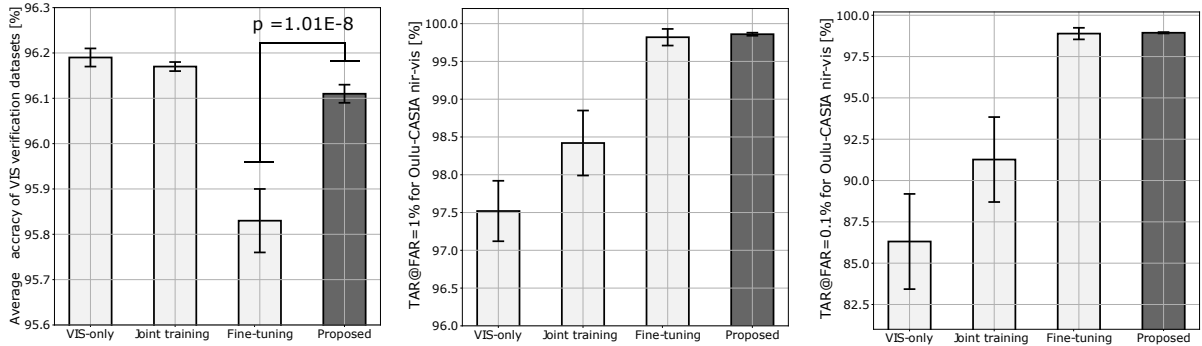}
   \end{center}
   \caption{Verification performances of three baseline models (VIS-only, joint training, and fine-tuning) and our proposed method. Left: Average accuracy of VIS verification datasets. Center and right: performances on Oulu-CAISA NIR\&VIS dataset at FAR = $1\%$ and FAR = $0.1\%$, respectively. Two-way ANOVA and Tukey-Kramer multi-comparison test are employed to evaluate p-value between the fine-tuning model and our proposed method.}
   \label{fig:results_1}
\end{figure*}

\subsection{Experimental settings}

\subsubsection{Datasets and evaluation protocols}

For the large-scale VIS dataset, we use MS1MV2 \cite{ArcF} which is a semi-automatic refined version of the MS-Celeb-1M dataset \cite{MS1M}.
There are about $5.8$ million VIS images of $85K$ identities in total.

For the small-size HFR dataset, we use the Oulu-CASIA NIR\&VIS dataset \cite{OuluCASIA} which contains both NIR and VIS face images in each identity.
There are $80$ identities with six expression variations. 
For a fair comparison with the previous studies, we have employed the same number of identities used in \cite{WCNN, DVR, DVG, DVGFace} ($20$ identities as the training set and $20$ identities as the testing set).
Further, eight face images from each expression are randomly selected from both NIR and VIS sets, which brings us a total of $96$ images per subject ($48$ NIR images and $48$ VIS images). 
In the verification phase, all the VIS images of the $20$ subjects are used as the gallery set and all the NIR images are treated as the probe set. 
Similarity scores are defined by the cosine similarity between the probe set and the gallery set.
We report TAR@FAR = $1\%$ and TAR@FAR = $0.1\%$ for comparisons.
Note that there are two widely used HFR face dataset, CASIA NIR-VIS 2.0 and BUAA VISNIR, other than Oulu-CASIA NIR\&VIS dataset. 
Unfortunately, the official download page for these datasets has not been found and we were unable to obtain them.

To evaluate the VIS performance, we conduct our experiments with six commonly used datasets: LFW \cite{LFW}, CALFW \cite{CALFW}, CPLFW \cite{CPLFW}, CFP (-FP, -FF) \cite{CFP}, and AgeDB-30 \cite{AgeDB30}. 
LFW contains $13,233$ images from $5,749$ identities and provides $6,000$ pairs from them. 
CALFW and CPLFW are the reorganized datasets from LFW to include higher pose and age variations.
CFP-FP and CFP-FF contain $500$ subjects, where the CFP-FP includes both frontal and profile images while the CFP-FF includes only the frontal images.
AgeDB-30 contains $12,240$ images of $440$ identities.
We compare the verification accuracy for identity pairs on these datasets.

All input face images are cropped to $112\times112$ pixels according to five facial points detected by MTCNN \cite{MTCNN} and normalized by subtracting 127.5 and then dividing by 128 for each pixel.
To augment the input data, we randomly flipped the input images horizontally.

\subsubsection{Baseline models and hyperparameter settings}

To determine the effectiveness of our method, we compare it with three baseline models: (1) a model trained with only the large-scale VIS dataset (VIS-only), (2) a model jointly trained with both the VIS dataset and the HFR dataset (joint training (JT)), and (3) a fine-tuned model from the VIS-only with the HFR dataset (fine-tuning (FT)).

Throughout the experiments, we utilize a single network architecture, LResNet50E \cite{ArcF}, which is a modified version of ResNet-50 \cite{ResNet}.
The angular-margin softmax loss \cite{CosF} is used for a loss function, with the margin parameter set to $m = 0.45$ and the scale parameter set to $s = 32$.
We train these models by the momentum SGD algorithm with the batch size of $128$.
The momentum and weight decay are set to $0.9$ and $5e-4$, respectively.

In the training of VIS-only and JT, the learning rate starts from $0.1$ and is divided by $10$ at $9$ and $14$ epoch, and the training process is finished at $16$ epochs.
For the training of our method, the batch sizes $N$ and $N^\prime$ are set to $128$.
The initial parameters $\Theta_{\mathrm{init}}$ are given from those of the VIS-only model.
The hyperparameters $\lambda$, $\alpha$ are set to $0.01$, $0.2$, respectively.
The learning rate is set to $0.001$ and the training process is finished at $100$ epochs.
The training of FT is conducted by setting $\alpha = 1.0$ in our method.

\subsection{Evaluation results}

In this section, we investigate the effectiveness of our method by comparing the evaluation results with the baseline models.
In order to obtain statistically reliable results, we conduct multiple trials of training for each model by randomly changing the initial variables and shuffling the order of input images, and evaluate the standard deviation for their verification performance.
We conduct the two-way analysis of variance (ANOVA) \cite{twoANOVA} with two factors, model (the proposed model vs. fine-tuning) and domain (VIS-VIS vs. NIR-VIS), followed by the Tukey-Kramer multi-comparison test \cite{TKcomp1,TKcomp2}.

First, we discuss the verification performances in VIS face images. 
The left panel of Fig.~\ref{fig:results_1} shows the average accuracy of six VIS verification datasets (LFW, CALFW, CPLFW, CFP-FP, CFP-FF, and AgeDB-30).
The accuracy of each dataset and the averaged values are also listed in Table~\ref{tab:results_1_vis}.
From the figure and the table, we can see the VIS-only model achieved the highest performances for all verification datasets.
The joint training model achieved comparable performances with the VIS-only model within the standard deviation.
In the fine-tuning model, we can observe a significant performance degradation compared to the VIS-only model, while in contrast, our proposed method could maintain the VIS performances in all verification datasets.
The performance degradation of our method ($0.08$ points in the averaged value) was significantly smaller ($77.8\%$) than that of the fine-tuning model ($0.36$ points in the averaged value).

Next, we discuss the HFR performances.
The center and right panels of Fig.~\ref{fig:results_1} shows the verification rate on the Oulu-CASIA NIR\&VIS dataset at $\mathrm{FAR} = 1\%$ and $\mathrm{FAR} = 0.1\%$, respectively.
The numerical values are also listed in Table~\ref{tab:results_1_HFR}.
From these results, we can see a slight improvement of the HFR performance in the joint training model compared to the VIS-only model.
The fine-tuning model achieved a higher performance than the joint training model by a large margin.
In addition, our method obtained a high performance comparable to that of the fine-tuning model.

It is worth noting that the standard deviations of both VIS and HFR performances in our proposed method are smaller than those of the joint training and fine-tuning model.
This observation could be understood from following facts.
First, the joint training model trains with large number of VIS face images and small number of HFR face images, in which the HFR performances have large deviations due to small appearances of the HFR images during a training.
Next, the fine-tuning model trains only with the HFR images, which causes large deviations of VIS performances.
In our proposed method, both the HFR and large-scale VIS face images are trained effectively so that the standard deviations of both HFR and VIS performacnes are being small.

\begin{table}
   \begin{center}
      \begin{tabular}{|l|c|c|}
         \hline
         Method & $\mathrm{FAR} = 1\%$ & $\mathrm{FAR} = 0.1\%$ \\
         \hline\hline
         VIS-only & 97.52 (0.40) & 86.31 (2.88) \\
         Joint training & 98.42 (0.43) & 91.27 (2.57) \\
         Fine-tuning & 99.82 (0.11) & 98.89 (0.35) \\
         Proposed & {\bf 99.86 (0.02)} & {\bf 98.94 (0.04)} \\
         \hline
         \end{tabular}
      \end{center}
   \caption{Comparison results with three baseline models (VIS-only, joint training, and fine-tuning) and our proposed method on the Oulu-CAISA NIR\&VIS dataset. Parentheses indicate standard deviations.}
   \label{tab:results_1_HFR}
\end{table}

\subsection{Feature distributions}

\begin{figure}[t]
   \begin{center}
      \includegraphics[width=0.7\linewidth]{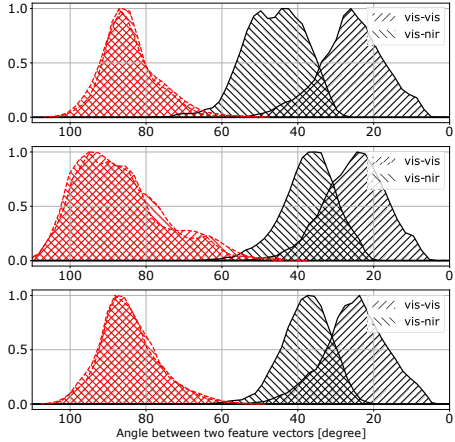}
   \end{center}
   \caption{Angle distributions of embedded features for NIR-VIS matching and VIS-VIS matching on the pre-trained model (upper), the fine-tuned model (center), and the model trained with our method (lower).
   Solid (black) and dashed (red) lines indicate genuine and imposter scores, respectively.
   All distributions are normalized so that the maximum value is 1.
   }
   \label{fig:feature_distrib}
\end{figure}

In this subsection, we demonstrate the effectiveness of our proposed JFDAL in terms of the distributions of embedded features.
To this end, we calculate the angle between embedded features which represents the relative relationship.
Figure~\ref{fig:feature_distrib} shows the angle distributions for NIR-VIS and VIS-VIS verification, where solid (black) and dashed (red) lines indicate the genuine and imposter scores, respectively.
All distributions are normalized so that the maximum value is 1 for easy comparison of distributions.
In the VIS-only model (upper panel of the figure), the distributions of genuine and imposter scores for the VIS-VIS pair are well separated.
In contrast, the genuine scores for the NIR-VIS pair were shifted to a large angle, which indicates a large domain gap.
The distributions for the fine-tuned model (center panel of the figure) shows that the fine-tuning could reduce the domain gap (i.e., genuine distributions of VIS-VIS and NIR-VIS are close).
In this case, however, the imposter distributions were highly collapsed and its tail expands.
Thus, the verification performances in the VIS face images were degraded in the fine-tuned model.
In contrast, our proposed JFDAL (lower panel of the figure) could simultaneously reduce the domain gap and retain the VIS feature distribution from the original one.

\subsection{Comparison to other HFR methods}

\begin{table}
   \begin{center}
      \begin{tabular}{|l|c|c|}
         \hline
         Method & $\mathrm{FAR} = 1\%$ & $\mathrm{FAR} = 0.1\%$ \\
         \hline\hline
         DVG \cite{DVG} & 98.5 & 92.9 \\
         SADG \cite{SADG} & 98.9 & 93.2 \\
         MMDL \cite{MMDL} & -- & 97.2 \\
         DVG-Face \cite{DVGFace} & 99.2 & 97.3 \\
         RGM \cite{RGM} & 99.69 & {\bf 98.96} \\ \hline
         Proposed & {\bf 99.86 (0.02)} & {\bf 98.94 (0.04)} \\
         \hline
         \end{tabular}
      \end{center}
   \caption{Comparison results with state-of-the-arts HFR methods on the Oulu-CAISA NIR\&VIS dataset. Parentheses indicate standard deviations.}
   \label{tab:comparison_sota}
\end{table}

In this subsection, we compare the verification performance of our proposed JFDAL on the Oulu-CASIA NIR\&VIS dataset with several state-of-the-arts methods: DVG \cite{DVG}, SADG \cite{SADG}, MMDL \cite{MMDL}, DVG-Face \cite{DVGFace}, and RGM \cite{RGM}.
Table~\ref{tab:comparison_sota} lists the results of the true acceptance ratio (TAR) at $\mathrm{FAR} = 1\%$ and $\mathrm{FAR} = 0.1\%$, where the numerical values of the conventional methods are referenced from their original papers.
From these results, we can see that our proposed JFDAL not only maintained the VIS performance but also achieved comparable performances to the state-of-the-art methods.
We also compare the performances with FAwF \cite{FAwF}, which is a state-of-the-art method on incremental learning for face recognition.
Since there are no evaluation results on Oulu-CASIA NIR\&VIS in \cite{FAwF}, we implement their method ourselves to determine the performances for the VIS and HFR datasets.
Consequently, we obtained $95.83 (0.07) \%$ for the average accuracy of six VIS verification datasets and $98.84 (0.05) \%$, $95.27 (0.08) \%$ for the verification rate of Oulu-CASIA NIR\&VIS at FAR$=1\%, 0.1\%$, respectively.
We found that our proposed JFDAL exceeded FAwF on both the VIS and HFR performances.

\subsection{Hyperparameter analysis} \label{sec:hparam_analysis}

\begin{figure}[t]
   \begin{center}
      \includegraphics[width=0.8\linewidth]{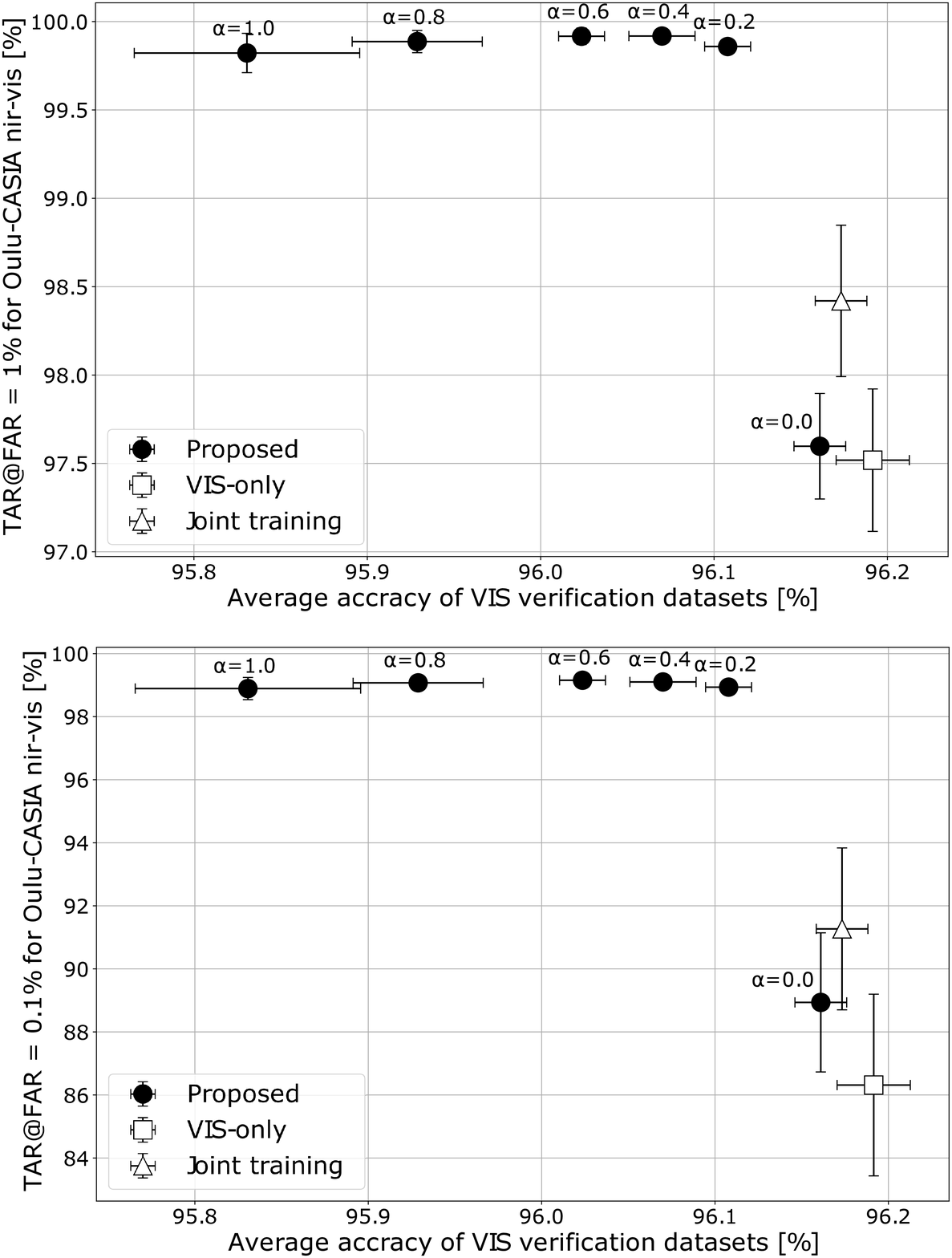}
   \end{center}
   \caption{Verification performances of proposed JFDAL with different $\alpha$ on the Oulu-CASIA NIR\&VIS against average accuracy of six VIS verification datasets.
   Standard deviations are evaluated for each model by conducting multiple trials of training with randomly changing the initial variables and shuffling the order of input images.}
   \label{fig:hparam_a}
\end{figure}

In this subsection, we conduct a hyperparameter search for the $\alpha$ that balance between CFDAL and SFDAL, as in Eq.~\ref{eq:Ltot}.
Figure~\ref{fig:hparam_a} shows the verification performance of the HFR dataset against the VIS dataset for several fixed $\alpha$.
For comparison, we also show the results of the two baseline models, VIS-only and joint training.
Note that the results of $\alpha=1.0$ are regarded as those of fine-tuning model.
From Fig.~\ref{fig:hparam_a}, we can see that the performances of the VIS dataset were gradually improved when decreasing the value of $\alpha$.
This observation indicates that the SFDAL component plays a significant role in maintaining the VIS performances.
Further, we find that the HFR performances were still high even in with a small $\alpha$.
For a very small $\alpha$, the HFR performances are suddenly dropped, in which an effect of CFDAL is too small to boost the HFR performacnes.
By taking into account both the VIS and HFR performance, we set the best value of $\alpha$ to $0.2$.

\section{Conclusions}

In this paper, we have tackled the challenging task of achieving a verification performance for heterogeneous face recognition (HFR) while maintaining the performance in visible (VIS) light spectrum face images.
We proposed a new learning framework, joint feature distribution alignment learning (JFDAL), that aligns feature distributions in different domains while simultaneously keeping the feature distributions for VIS face images with their original distributions.
As a result, the domain gap is effectively reduced without significant collapse of the VIS distributions, which occurs when conventional fine-tuning is used.
The results of extensive experiments demonstrate the effectiveness of our method compared to conventional fine-tuning and joint training.
We also showed that our method can achieve state-of-the-art HFR performances.

Since our method is not specific to VIS-NIR domain, it could be applicable to any domain as long as we can define the difference of domain in data/feature level (e.g., pose variations).
We will continue to explore more applications in our future work.

\section*{Acknowledgements}
The authors thank the anonymous reviewers for their careful reading to improve the manuscript. We would also like to thank Koichi Takahashi, Makoto Takamoto, Hiroshi Fukui, Ryosuke Sakai and Tomokazu Kaneko for the insightful discussions and supports of the project.


{\small
\begin{thebibliography}{99}
   \bibitem{DVGFace}
   Chaoyou Fu, Xiang Wu, Yibo Hu, Huaibo Huang, and Ran He. DVG-Face: Dual Variational Generation for Heterogeneous Face Recognition. In {\it TPAMI}, 2020.
   \bibitem{MMD}
   Arthur Gretton, Karsten Borgwardt, Malte J. Rasch, Bernhard Scholkopf, and Alexander J. Smola. A Kernel Method for the Two-Sample-Problem. In {\it NIPS}, 2006.
   \bibitem{MMDcnn}
   E. Tzeng, J. Hoffman, N. Zhang, K. Saenko, and T. Darrell. Deep domain confusion: Maximizing for domain invariance. In {\it arXiv:1412.3474}, 2014.
   \bibitem{LeNet}
   Y. LeCun, L. Bottou, Y. Bengio, and P. Haffner. Gradient-based learning applied to document recognition. In {\it Proceedings of the IEEE, 86. 11}, 1998.
   \bibitem{MS1M}
   Y. Guo, L. Zhang, Y. Hu, X. He, and J. Gao. Ms-celeb-1m: A dataset and benchmark for large-scale face recognition. In {\it ECCV. Springer, 2016, pp. 87–102}, 2016.
   \bibitem{HFRsurvey}
   S. Ouyang, T. Hospedales, Y.-Z. Song, X. Li, C. C. Loy, and X. Wang. A survey on heterogeneous face recognition: Sketch, infra-red, 3d and low-resolution, In {\it Image and Vision Computing, vol. 56, pp. 28–48}, 2016.
   \bibitem{Fsynth_1}
   He Zhang, Benjamin S. Riggan, Shuowen Hu, Nathaniel J. Short, and Vishal M. Patel. Synthesis of high-quality visible faces from polarimetric thermal faces using generative adversarial networks. In {\it IJCV}, 2019.
   \bibitem{Fsynth_2}
   Lingxiao Song, Man Zhang, Xiang Wu, and Ran He. Adversarial discriminative heterogeneous face
recognition. In {\it AAAI}, 2018.
   \bibitem{Fsynth_3}
   Boyan Duan, Chaoyou Fu, Yi Li, Xingguang Song, and Ran He. Cross-Spectral Face Hallucination via Disentangling Independent Factors. In {\it CVPR}, 2020.
   \bibitem{DinvF_1}
   Cedric Nimpa Fondje, Shuowen Hu, Nathaniel J. Short, and Benjamin S. Riggan. Cross-Domain Identification for Thermal-to-Visible Face Recognition. In {\it IJCB}, 2020.
   \bibitem{WCNN}
   R. He, X. Wu, Z. Sun, and T. Tan. Wasserstein CNN: Learning invariant features for nir-vis face recognition. In {\it TPAMI, vol. 41, no. 7, pp. 1761–1773}, 2018.
   \bibitem{MMDL}
   Bing Cao, Nannan Wang, Xinbo Gao, Jie Li, and Zhifeng Li. Multi-Margin based Decorrelation Learning for Heterogeneous Face Recognition. In {\it IJCAI}, 2019.
   \bibitem{DVR}
   X. Wu, H. Huang, V. M. Patel, R. He, and Z. Sun. Disentangled variational representation for heterogeneous face recognition. In {\it AAAI}, 2019.
   \bibitem{PrjComm_1}
   Wang, K., He, R., Wang, W., Wang, L., and Tan, T. Learning coupled feature spaces for cross-modal matching. In {\it ICCV}, 2013.
   \bibitem{PrjComm_2}
   Yi, D., Lei, Z., Liao, S., and Li, S. Shared representation learning for heterogeneous face recognition. In {\it FG}, 2015.
   \bibitem{PrjComm_3}
   Meina Kan, Shiguang Shan, Haihong Zhang, Shihong Lao, and Xilin Chen. Multi-view discriminant analysis. In {\it TPAMI, 38(1):188–194}, 2016.
   \bibitem{PrjComm_4}
   Z. Li, D. Gong, Q. Li, D. Tao, and X. Li. Mutual component analysis for heterogeneous face recognition. In {\it ACM Transactions on Intelligent Systems and Technology, vol. 7, no. 3, pp. 1–23}, 2016
   \bibitem{VGG2}
   Q. Cao, L. Shen, W. Xie, O. M. Parkhi, and A. Zisserman. VGGFace2: A dataset for recognising faces across pose and age. In {\it FG}, 2018.
   \bibitem{ArcF}
   J. Deng, J. Guo, N. Xue, and S. Zafeiriou. ArcFace: Additive Angular Margin Loss for Deep Face Recognition. In {\it CVPR}, 2019.
   \bibitem{DVG}
   C. Fu, X. Wu, Y. Hu, H. Huang, and R. He. Dual variational generation for low shot heterogeneous face recognition. In {\it NeurIPS}, 2019.
   \bibitem{CFsurvey}
   G. I. Parisi, R. Kemker, J. L. Part, C. Kanan, and S. Wermter. Continual Lifelong Learning with Neural Networks: A Review. In {\it Neural Networks, 113, 54}, 2019.
   \bibitem{LFW}
   G. B. Huang, M. Ramesh, T. Berg, and E. Learned-Miller. Labeled faces in the wild: A database for studying face recognition in unconstrained environments. Technical report, 2007.
   \bibitem{MultiTask}
   R. Caruana. Multitask learning. In {\it Machine learning, 28, 41}, 1997.
   \bibitem{CFP}
   S. Sengupta, J.-C. Chen, C. Castillo, V. M. Patel, R. Chellappa, and D. W. Jacobs. Frontal to profile face verification in the wild. In {\it WACV}, 2016.
   \bibitem{AgeDB30}
   S. Moschoglou, A. Papaioannou, C. Sagonas, J. Deng, I. Kotsia, and S. Zafeiriou. Agedb: The first manually collected in-the-wild age database. In {\it CVPR Workshop}, 2017.
   \bibitem{CASIA}
   D. Yi, Z. Lei, S. Liao, and S. Z. Li. Learning face representation from scratch. In {\it arXiv:1411.7923}, 2014
   \bibitem{MTCNN}
   K. Zhang, Z. Zhang, Z. Li, and Y. Qiao. Joint Face Detection and Alignment using Multi-task Cascaded Convolutional Networks. In {\it Signal Processing Letters, 23. 10}, 2016.
   \bibitem{ResNet}
   K. He, X. Zhang, S. Ren, and J. Sun. Deep residual learning for image recognition. In {\it CVPR}, 2016
   \bibitem{OuluCASIA}
   J. Chen, D. Yi, J. Yang, G. Zhao, S. Z. Li, and M. Pietikainen. Learning mappings for face synthesis from near infrared to visual light images. In {\it CVPR}, 2009
   \bibitem{CosF}
   H. Wang, Y. Wang, Z. Zhou, X. Ji, Z. Li, D. Gong, J. Zhou, and W. Liu. Cosface: Large margin cosine loss for deep face recognition. In {\it CVPR}, 2018.
   \bibitem{LwF}
   Z. Li and D. Hoiem. Learning without forgetting. In {\it TPAMI, vol. 40, no. 12, pp. 2935–2947}, 2017.
   \bibitem{Distil}
   G. Hinton, O. Vinyals, and J. Dean. Distilling the knowledge in a neural network. In {\it arXiv:1503.02531}.
   \bibitem{iCaRL}
   S.-A. Rebuffi, A. Kolesnikov, G. Sperl, and C. H. Lampert. iCaRL: Incremental classifier and representation learning. In {\it CVPR, pp. 2001–2010}, 2017.
   \bibitem{FAwF}
   Hao Liu, Xiangyu Zhu, Zhen Lei, Dong Cao, and Stan Z. Li. Fast Adapting without Forgetting for Face Recognition. In {\it IEEE Transactions on Circuits and Systems for Video Technology}, 2020.
   \bibitem{CALFW}
   T. Zheng, W. Deng, and J. Hu. Cross-age LFW: A database for studying cross-age face recognition in unconstrained environments. In {\it arXiv:1708.08197}, 2017.
   \bibitem{CPLFW}
   T. Zheng and W. Deng. Cross-pose LFW: A database for studying cross-pose face recognition in unconstrained environments. In {\it Beijing University of Posts and Telecommunications, Technical Report 18-01, February}, 2018.
   \bibitem{RGM}
   MyeongAh Cho, Taeoh Kim, Ig-Jae Kim, Kyungjae Lee, Sangyoun Lee. Relational Deep Feature Learning for Heterogeneous Face Recognition. In {\it IEEE Transactions on Information Forensics and Security, vol. 16, pp. 376-388}, 2021.
   \bibitem{SADG}
   Rushuang Xu, MyeongAh Cho, and Sangyoun Lee. A NIR-to-VIS face recognition via part adaptive and relation attention module. In {\it arXiv:2102.00689 [cs.CV]}.
   \bibitem{twoANOVA}
   Fisher R.A. Statistical methods for research workers. 1925.
   \bibitem{TKcomp1}
   John W. Tukey. Comparing individual means in the analysis of variance. In {\it Biometrics vol. 5, no. 2, pp. 99-114}, 1949.
   \bibitem{TKcomp2}
   Kramer Clyde Young. Extension of multiple range tests to group means with unequal numbers of replications. In {\it Biometrics vol. 12, no. 3, pp. 307-310}, 1956.
   \bibitem{TeaNet}
   Duong, C. N., Luu, K., Quach, K. G. and Le, N. ShrinkTeaNet: Million-scale lightweight face recognition via shrinking teacher-student networks. In {\it arXiv:1905.10620}.
\end{thebibliography}
}
\end{document}